\documentclass[]{style/style}

\usepackage{microtype}
\usepackage{graphicx}
\usepackage{subcaption}
\usepackage{csquotes}
\usepackage{afterpage}
\usepackage{xcolor}

\usepackage{amsmath}
\usepackage{amssymb}
\usepackage{mathtools}
\usepackage{amsthm}
\usepackage{xspace}
\usepackage{colortbl}
\usepackage{multirow}
\usepackage{multicol}
\usepackage{enumitem}
\usepackage{wrapfig}
\usepackage{nicefrac}
\usepackage[font=small,labelfont=bf,justification=justified,singlelinecheck=false]{caption}

\usepackage{booktabs}
\usepackage{multirow}
\usepackage{graphicx}
\usepackage{xcolor}
\usepackage{geometry}
\usepackage{array}
\usepackage{tabularx}
\newcolumntype{L}[1]{>{\raggedright\arraybackslash}p{#1}}
\newcolumntype{Y}{>{\raggedright\arraybackslash}X}

\usepackage{amsmath}

\title{Wan-Streamer v0.2: Higher Resolution, Same~Latency}

\author[]{Wan Team, Alibaba Group}

\affiliation[]{\small See \hyperref[app:contributions-and-acknowledgements]{Contributions and Acknowledgements} for the full author list.}

\abstract{
We present \textbf{Wan-Streamer v0.2}, a latency-preserving upgrade of the native-streaming, end-to-end audio-visual interaction model. v0.2 keeps the v0.1 modeling formulation, but raises the interactive output stream from 192$\times$336 to \textbf{640$\times$368} while preserving approximately \textbf{200 ms} model-side signal-to-signal latency at 25 FPS. The higher-resolution stream supports \textbf{scene-grounded mid-shot agents} whose posture, gaze, hands, nearby objects, and local scene layout remain legible during real-time conversation. To support the larger visual stream without adding user-visible delay, v0.2 keeps the thinker as a single-GPU low-latency path for streaming perception, the short language/state Transformer pass that builds the generation cache, and final decoding. The performer becomes a multi-GPU Ulysses-style context-parallel group for the expensive next-unit latent generation. Each performer rank writes incoming K/V into a pre-sharded local cache. The long high-resolution latent video sequence is split across ranks for denoising and gathered through Ulysses communication, while the much shorter audio latent sequence is generated without sequence sharding. In this split, the thinker's language/state computation reaches the performer only as K/V conditioning, so no separate language sequence has to be communicated inside the performer group. This concentrates additional hardware on visual generation while preserving the compact thinker-performer boundary, keeping total remote interaction latency at approximately 550 ms when a 350 ms bidirectional network budget is included.
}

\metadata[Website]{\url{https://wan-streamer.com/}}

\begin{document}

\maketitle

\section{Introduction}
\label{section:intro}

Wan-Streamer v0.2 directly upgrades Wan-Streamer v0.1~\citep{huang2026wanstreamerv01endtoendrealtime}. Real-time audio-visual interaction sits at the intersection of full-duplex spoken dialogue, multimodal perception, streaming video generation, and interactive digital humans. Full-duplex speech systems show that natural dialogue should not be reduced to alternating ASR--LLM--TTS turns~\citep{chen2025fullduplexsurvey,defossez2024moshi}. Omni-modal models extend perception to image, video, and audio inputs~\citep{qwen3omni2025,minicpmo45}, while video generation and causal rollout methods provide the visual synthesis and streaming foundations needed for interactive output~\citep{wan2025wan,chen2024diffusionforcing,huang2025selfforcing}. In parallel, real-time avatars and digital-human systems have advanced audio-driven faces, streaming visual agents, and end-to-end embodied interaction~\citep{xu2024vasa,streamavatar2025,mavid2025,ao2024bodyofher}.

Wan-Streamer v0.1 established a native-streaming formulation for this setting: user and agent text, audio, and video are represented on one causal timeline and modeled by a single Transformer. Unlike cascaded visual-agent systems, this formulation keeps perception, response timing, speech, visible listening behavior, and synchronized video response inside one causal interaction state. It closes the audio-visual interaction loop, but the preliminary 192p output limits the visual range. Close-up video-call framing preserves facial response and speaking behavior, while wider compositions leave body posture, nearby objects, and scene context too compressed for scene-grounded interaction.

Wan-Streamer v0.2 is a latency-preserving resolution upgrade. It raises the interactive output stream from 192$\times$336 to 640$\times$368 at 25 FPS while keeping approximately 200 ms model-side response latency. This target is constrained by streaming causality rather than offline rendering quality: every 160 ms unit must process current user observations, update the shared interaction state, generate synchronized speech and video latents, decode the previous unit, and emit the response without stretching the interaction cadence. With a 350 ms bidirectional network budget, the resulting remote interaction latency remains approximately 550 ms.

The larger stream changes the usable visual composition. v0.2 improves close-up video-call fidelity and supports scene-grounded mid-shot agents whose posture, gaze, hands, nearby objects, and local scene layout remain legible during real-time conversation. This expands the visual format from portrait-like calls toward situated conversations in which the agent remains visibly grounded in its surroundings.

To meet the same latency budget, v0.2 changes the serving topology while keeping the native-streaming formulation fixed. The thinker remains a single-GPU, low-latency path for streaming perception, language/state update, KV-cache construction, and final causal decoding; this language/state pass is the part that produces the K/V cache used by generation. The 640$\times$368 latent generation path moves into a multi-GPU Ulysses-style context-parallel performer group~\citep{jacobs2023deepspeedulysses}. The thinker broadcasts compact performer-compatible K/V slices, each performer rank writes them into its pre-sharded local cache, and denoising for the long latent video sequence is split across ranks with Ulysses all-to-all/gather communication. Audio latents for each streaming unit contain far fewer tokens, so they are generated without sequence sharding. This isolates the added visual-generation cost from the latency-critical control path while preserving the compact thinker-performer boundary.

\begin{figure}[t]
    \centering
    \includegraphics[width=\linewidth]{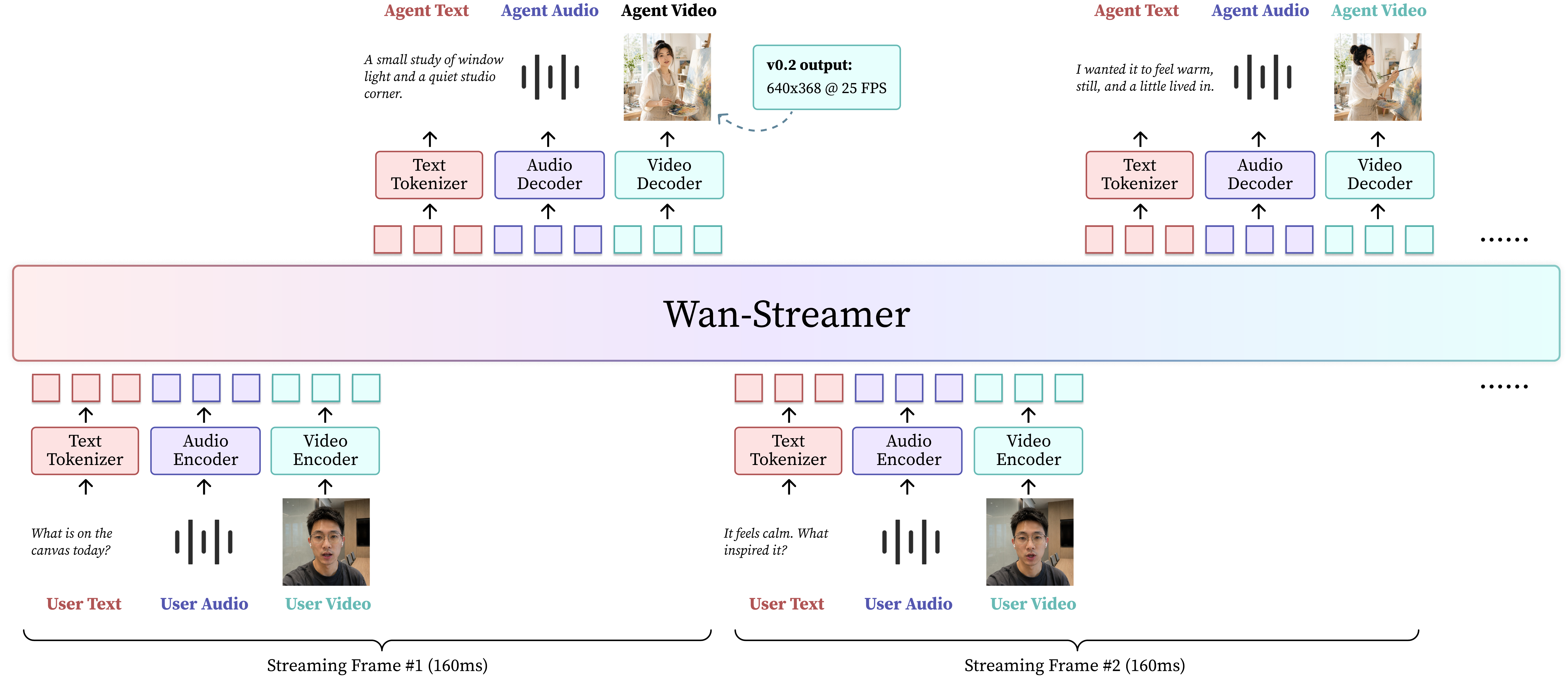}
    \caption{Wan-Streamer remains one native-streaming model: language, audio, and video inputs and outputs are represented on a shared causal timeline and coordinated by block-causal attention. v0.2 keeps this formulation while increasing the output resolution and changing the deployment strategy described in Fig.~\ref{fig:thinker-performer-overlap}.}
    \label{fig:framework}
\end{figure}

Figure~\ref{fig:framework} shows the native-streaming formulation inherited from v0.1. Against this baseline, v0.2 changes three axes: the output stream increases to 640$\times$368, the high-cost latent generation path moves into a Ulysses-style context-parallel performer, and the supported visual composition expands from close-up calls to scene-grounded mid-shot agents. The rest of the paper follows these axes: Sec.~\ref{sec:upgrade} summarizes the version comparison, Sec.~\ref{sec:serving} describes latency-preserving serving, and Sec.~\ref{sec:experiments} describes the experiments.

Our contributions are:
\begin{itemize}[leftmargin=1.3em]
    \item We upgrade Wan-Streamer from 192$\times$336 to 640$\times$368 video output while keeping approximately 200 ms model-side response latency.
    \item We introduce a v0.2 serving topology with a single-GPU thinker and a Ulysses-style context-parallel multi-GPU performer, using pre-sharded performer-side K/V caches and sequence parallelism for the high-resolution latent video denoising path.
    \item We expand the visual interaction scope from close-up calls to higher-fidelity close-up interactions and scene-grounded mid-shot agents with readable body and scene context.
\end{itemize}

\FloatBarrier

\section{Upgrade Design}
\label{sec:upgrade}

\subsection{Stable native-streaming formulation}
v0.2 keeps the core Wan-Streamer formulation stable. The model is still trained as one end-to-end causal stream: user text, audio, and video observations update the same history that conditions agent text, speech, and video responses. Generated audio-video latents are committed back into history after each unit, so the next response can depend on both the user's recent behavior and the agent's own previous expression. This formulation is inherited from v0.1 and serves as the baseline for the v0.2 upgrade.

v0.2 expands the visual range around this formulation. The visual target moves from 192p to 640$\times$368, and the data emphasis moves from mostly close-up call framing to wider, scene-grounded conversational settings. In mid-shot composition, the model must preserve identity, gaze, hand and torso posture, local objects, and scene layout while continuing to listen and speak in real time.

\subsection{Version comparison}
Table~\ref{tab:v02-upgrade} summarizes the version-level changes in v0.2: the end-to-end streaming formulation and latency budget stay fixed, while output resolution, visual format, and serving topology change.

\begin{table}[!ht]
    \caption{Summary of the Wan-Streamer v0.1 to v0.2 upgrade. Emphasized cells indicate changed components or preserved latency targets.}
    \label{tab:v02-upgrade}
    \centering
    \footnotesize
    \setlength{\tabcolsep}{5pt}
    \renewcommand{\arraystretch}{1.16}
    \begin{tabularx}{\textwidth}{@{}L{.20\textwidth}L{.33\textwidth}Y@{}}
        \toprule
        Aspect & v0.1 & v0.2 \\
        \midrule
        Output resolution & 192$\times$336 & \textbf{640$\times$368} \\
        Frame rate & 25 FPS & 25 FPS \\
        Model-side latency & $\sim$200 ms & $\sim$200 ms, \textbf{unchanged} \\
        Total interaction latency & $\sim$550 ms with a 350 ms bidirectional network budget & $\sim$550 ms with the same 350 ms bidirectional network budget \\
        Thinker & Streaming perception, state update, KV construction, and decoding & Same role, \textbf{kept on one GPU} \\
        Performer & Single-GPU latent generation & \textbf{Multi-GPU Ulysses-style context-parallel} latent generation \\
        Communication & Thinker-performer K/V and latent exchange & Thinker broadcasts performer-compatible K/V slices; Ulysses all-to-all/gather for the latent video sequence stays inside the performer group \\
        Visual presence & Close-up video-call framing & Higher-fidelity close-up interactions plus \textbf{scene-grounded mid-shot agents with readable body and scene context} \\
        \bottomrule
    \end{tabularx}
\end{table}

\FloatBarrier

\section{Latency-Preserving Serving}
\label{sec:serving}

The serving challenge in v0.2 is to allocate the additional 640$\times$368 generation cost without slowing down the interactive loop. We split the deployed model into two roles:

\begin{itemize}[leftmargin=1.3em]
    \item \textbf{Thinker.} A single GPU hosts the causal audio/video encoders, the token-causal Transformer path for language and state update, KV-cache construction, and the causal decoders that turn returned latents into output audio and video. The language/state path is reflected in the K/V cache that conditions generation.
    \item \textbf{Performer.} A Ulysses-style context-parallel GPU group hosts the expensive flow-matching latent generation path. Performer ranks keep pre-sharded K/V caches, split the long latent video sequence across ranks, and communicate through Ulysses all-to-all/gather collectives around attention. The audio latents are short enough that sequence sharding would add overhead rather than useful parallelism, so they are generated without sequence sharding.
\end{itemize}

\begin{figure}[t]
    \centering
    \includegraphics[width=\linewidth]{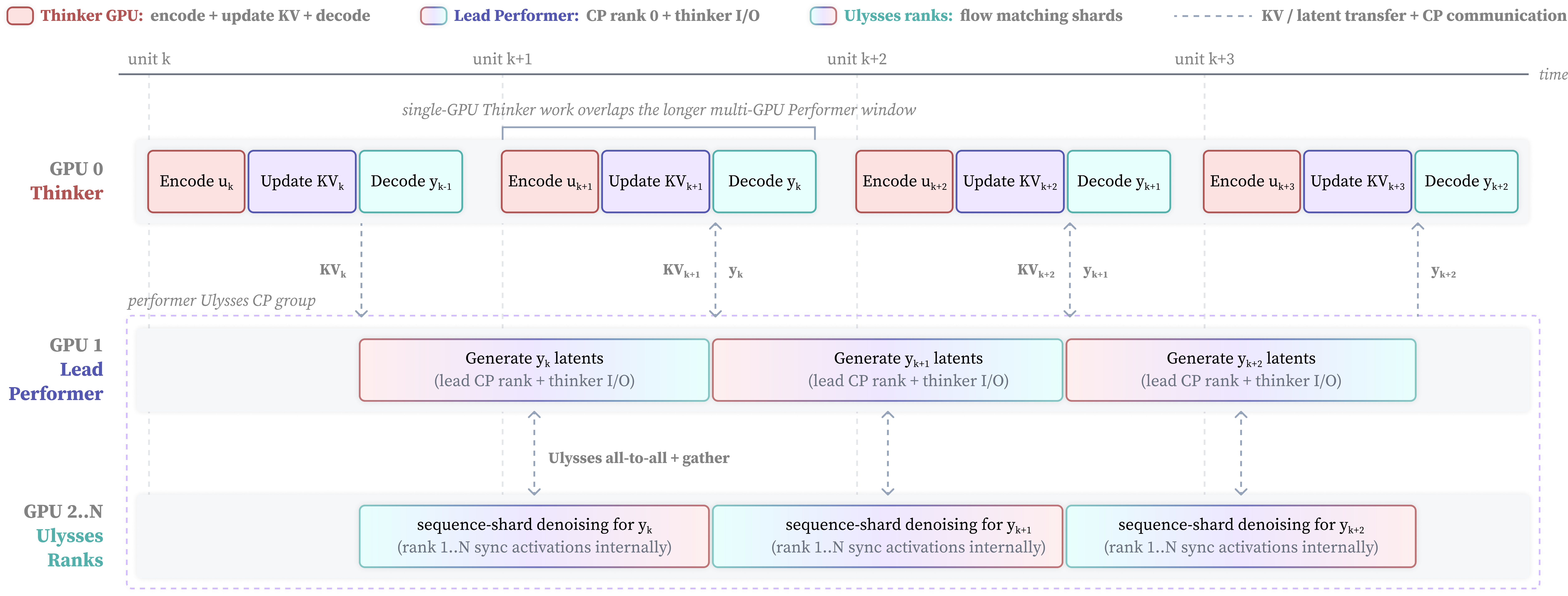}
    \caption{v0.2 latency-preserving serving. The thinker stays on one GPU and owns perception, language/state update, K/V construction, and final decoding. The performer uses Ulysses-style context parallelism for the 640$\times$368 latent generation path: K/V slices are written into pre-sharded performer caches, the high-resolution latent video sequence is split across ranks for denoising, and Ulysses all-to-all/gather communication stays inside the performer group. Audio latents are short and are not split across ranks.}
    \label{fig:thinker-performer-overlap}
\end{figure}

As shown in Fig.~\ref{fig:thinker-performer-overlap}, at streaming unit $k$, the thinker consumes the current user observations and produces a new performer-compatible K/V slice. Around the same boundary, it receives the previous unit's generated latents, decodes them, and emits the response. The performer ranks receive the current K/V slice, update their local shards of the full-history cache, and run Ulysses context-parallel denoising for the next unit. The high-resolution latent video is the main sequence-parallel path. The audio latents are much shorter, so they stay unsharded; the language/state computation has already been folded into the K/V slice produced by the thinker.

This schedule separates throughput from response latency. Real-time throughput requires the performer group time plus thinker-performer K/V and latent transfer plus intra-performer Ulysses communication to fit inside one 160 ms unit. The model-side response latency is the signal-to-signal path through encoding, state update, latent generation, and decoding; this remains approximately 200 ms. The key constraint is that the additional v0.2 work is concentrated in the context-parallel performer, while the thinker remains a compact low-latency interaction path.

\FloatBarrier

\section{Experiments}
\label{sec:experiments}

\textbf{Latency and runtime protocol.}
We use the same response boundary as Wan-Streamer v0.1. Model-side signal-to-signal latency starts when a 160 ms user streaming unit is available to the thinker and ends when the corresponding audio-video response unit has been decoded for emission. Under the serving path in Sec.~\ref{sec:serving}, v0.2 keeps approximately 200 ms model-side latency while producing 640$\times$368 video at 25 FPS. With the same 350 ms bidirectional network budget used in v0.1, the total remote interaction latency remains approximately 550 ms. We keep this network term as an external deployment assumption, so the comparison isolates the v0.2 model-side and serving changes; bandwidth-limited transport effects are outside the model-side latency measurement reported here.

Public real-time systems report different endpoints, including first-packet speech latency, first-frame delay, FPS, audio-to-visual delay, or product-level response time~\citep{doubao_realtime_voice2025,openai2024gpt4o,hume2025evi3,streamavatar2025,hallolive2026}. We therefore keep the v0.1 measurement convention and report the v0.2 runtime at the same response boundary.

\textbf{Qualitative visual observations.}
We use generated 640$\times$368 conversations as qualitative observations of the upgraded output format. The inspection focuses on visual stability and legibility during both listening and speaking intervals, including facial detail, gaze, mouth motion, hands, posture, nearby objects, and local scene layout.

These observations characterize the v0.2 output format: clearer close-up calls and scene-grounded mid-shot agents under the same low-latency streaming setting.

\section{Conclusion}

Wan-Streamer v0.2 keeps the native full-duplex formulation of v0.1 while raising the interactive stream from 192$\times$336 to 640$\times$368 at approximately 200 ms model-side latency. The single-GPU thinker preserves the latency-critical loop, while the Ulysses-style context-parallel performer absorbs the added visual latent-generation cost through pre-sharded K/V caches and sequence parallelism for the high-resolution latent video denoising path. This yields clearer video-call interaction and scene-grounded mid-shot agents under the same low-latency streaming setting.

\clearpage
\bibliographystyle{unsrtnat}
\bibliography{paper}

\clearpage

\appendix
\section*{Appendix}

\section{Contributions and Acknowledgements}
\label{app:contributions-and-acknowledgements}

\subsection{Core Contributors}
Lianghua Huang, Zhi-Fan Wu, Yupeng Shi, Wei Wang, Mengyang Feng, Junjie He, Chen-Wei Xie, Yu Liu, and Jingren Zhou.

\subsection{Contributors}
Contributors are listed alphabetically by first name: Ang Wang, Bang Zhang, Baole Ai, Chen Liang, Cheng Yu, Chongyang Zhong, Jinwei Qi, Kai Zhu, Pandeng Li, Peng Zhang, Wenyuan Zhang, Xinhua Cheng, Yitong Huang, Yun Zheng, Yuxiang Bao, Yuzheng Wang, and Zoubin Bi.

\end{document}